\documentclass[conference]{IEEEtran}

\usepackage{cite}
\usepackage{amsmath,amssymb,amsfonts}
\usepackage{algorithm}
\usepackage{algorithmicx} %

\usepackage[ruled,vlined, algo2e]{algorithm2e} 

\usepackage{graphicx}
\usepackage{subcaption}
\usepackage{textcomp}
\usepackage{xcolor}
\def\BibTeX{{\rm B\kern-.05em{\sc i\kern-.025em b}\kern-.08em
    T\kern-.1667em\lower.7ex\hbox{E}\kern-.125emX}}
\SetKwBlock{DummyBlock}{}{}
\begin{document}

\title{Automated Graph Genetic Algorithm based Puzzle Validation for Faster Game Design
}

\author{\IEEEauthorblockN{Karine Levonyan, Jesse Harder, Fernando De Mesentier Silva}
\IEEEauthorblockA{
\textit{EADP Data and AI, Electronic Arts,
Redwood Shores, CA, USA} \\
karine@ea.com, jharder@ea.com, fdemesentiersilva@ea.com}
}
\maketitle
\begin{abstract}
 Many games are reliant on creating new and engaging content constantly to maintain the interest of their player-base. One such example are puzzle games, in such it is common to have a recurrent need to create new puzzles. Creating new puzzles requires guaranteeing that they are solvable and interesting to players, both of which require significant time from the designers. Automatic validation of puzzles provides designers with a significant time saving and potential boost in quality. Automation allows puzzle designers to estimate different properties, increase the variety of constraints, and even personalize puzzles to specific players. Puzzles often have a large design space, which renders exhaustive search approaches infeasible, if they require significant time. Specifically, those puzzles can be formulated as quadratic combinatorial optimization problems. This paper presents an evolutionary algorithm, empowered by expert-knowledge informed heuristics, for solving logical puzzles in video games efficiently, leading to  a more efficient design process. We discuss multiple variations of hybrid genetic approaches for constraint satisfaction problems that allow us to find a diverse set of near-optimal solutions for puzzles.  We demonstrate our approach on a fantasy Party Building Puzzle game, and discuss how it can be applied more broadly to other puzzles to guide designers in their creative process.

\end{abstract}

\section{Introduction}
Puzzles have long been a staple of video games. 
They can be just enjoyable to play, or necessary to advance in the game, and usually provide in-game rewards upon their completion. In terms of design, puzzles can affect in-game economy, if completion of harder puzzles can lead to better rewards, and rewards can be purchased with, or traded for, in-game currency. Unlike  traditional logic puzzles like Sudoku, World Wheel, Tower of Hanoi, Refraction, crosswords etc, where game rules are predefined and static~\cite{puzzlenpcomplete}, puzzles in video games commonly have changing constraints in order to diversify gameplay. With many games having ever evolving puzzles which continue to grow in complexity and difficulty, the task of designing these puzzles becomes time consuming and challenging.

The game design process is usually complex and labor intensive. Exploring the design space is how game creators discover and develop the rules and mechanics of the game. One of the principal challenges in designing puzzles, and potentially the most important for game designers, is guaranteeing they can be solved by players~\cite{Smith_quantifyingover}. Unsolvable puzzles are frustrating to players for obvious reasons, but puzzles that become exceptionally hard due to virtually infeasible constraints, such as having barely enough time to perform all necessary moves, can be just as equally frustrating. In order to guarantee the quality of the player experience, designers then need to certify that solution for the puzzle exists. 
With ever growing constraints and numerous puzzles to create, the task quickly becomes more complex. 
In addition to time saving, by automating certain aspects of the puzzle creation process, it can also assist in discovering new solutions that might not have been noticed initially. This allows designers to validate and quickly iterate to different versions of their puzzles. It also answers questions such as: how many solutions exist for a particular puzzle, what is the optimal solution, or what is the ``cheapest price'' solution.

For the game showcased in this paper, game designers need to produce a significant amount of new puzzles everyday. In order for our system to be of assistance to their workflow, it is necessary that it is able to find solutions within a time frame that allows designers to make quick iterations of their design. This scenario and the sheer size of the search space of potential solutions renders straight-forward techniques, such as exhaustive search, impractical. 
In order to meet such requirements we propose an evolutionary algorithm to search the space of candidates, making use of an expert knowledge derived heuristic to guide exploration of potential puzzle solutions. Although this technique does not guarantee we will find the optimal solution, which is also of interest to the designers, it consistently finds diverse, close-to-optimal solutions under our strict time constraints. The diverse set of solutions found provides valuable insight to the designers allowing them to quickly analyze  the attributes of the solution space and compare different design iterations.  

The puzzles discussed in this paper are deterministic, single-player, and fully observable at all steps. These puzzles are defined as constraint satisfaction problems with one or more constraints, where the objective is to select and assign items, from a pool of candidates, in order to complete the requirements. If the puzzle is solvable, there is rarely (if ever) a unique solution to it.  In this paper, we describe a randomized heuristic-driven construction state-space search based methodology to validate the solvability of a puzzle. With thousands of possible combinations to build a solution from, the search efficiency is key. In addition, each solution also has `solution cost', which is the sum of individual item prices comprising the solution. The cost of the solution should be minimized to better assess the `value' of the puzzle offered to the players. 
We further expand this methodology to find a set of near-optimal solutions efficiently using a custom designed evolutionary algorithm. Important to emphasize, in our work we focus solely on puzzle validation, regardless if the puzzles were generated manually or automatically.

The rest of the paper is formatted as such: Section \ref{sec:relatedwork} discusses how our approach fits in the context of previous research. Section~\ref{PBP} defines the party building puzzle, our example game, followed by the constructive approach to build a solution~\ref{sec:formationplaytesting}. Section~\ref{sec:optimization} further develops our approach on how to harness power of a Genetic Algorithm not only find a solution but how to optimize the search for puzzle solutions, while trying to approximate the global price optimum. Lastly, we conclude by explaining our findings and future work in Section~\ref{sec:conclusionfuture}.

\section{Related Work}\label{sec:relatedwork}
The goal of this work is to develop an automated solution strategy targeted at validating a puzzle design, which lends itself well to the concept of automating playtesting. Commonly, automation in playtesting revolves around agents that can play through the game, or particular game scenarios. In particular, artificial intelligence agents could be used to test the boundaries of the rules/design constraints as in~\cite{de2017aicontemporaryboardgames, garcia2018automatedhearthstone}. While our approach is not conducted by an agent (puzzles are action-based), but rather the player needs to find a solution by selecting and organizing items that can fulfill the requirements presented.

Important to note that for our problem 
rather than automatically trying to change the design, or to create a new puzzle, we instead assist in evaluating the feasibility of a proposed puzzle. Perhaps more similar to our approach, Bhatt et al. evolve Hearthstone decks, by selecting from existing cards, to fit the strategy of game playing agents~\cite{bhatt2018exploringhearthstone}.

The strategy of using custom Genetic Algorithm based solvers for Jigsaw puzzles was first introduced in~\cite{Toyama2002AssemblyOP} for binary image puzzles. The large set of selection items presents a unique challenge in choosing the solution optimization approach. The closest example of application of an evolutionary approach to effectively solve puzzles rather than applying greedy approach is demonstrated by~\cite{SolomonPuzzleEvolution} for very large Jigsaw puzzles, and later extended to multiple puzzles by \cite{MultipleJigsaw}. Specifically, in~\cite{SolomonPuzzleEvolution} authors  propose an approach that iteratively improves initial population via the means of natural selection (mutation, selection, crossover)  to find more accurate solutions (i.e. the correct image) with the novel puzzle representation and a custom crossover approach.

The puzzles considered in our work have similarly large pool of items to (pre)select and assign, which poses comparable computation hurdle. The video game puzzles we consider pose an added challenge of non-uniqueness of a solution as well as absence of a clearly defined benchmark solution to compare with.

\section{Example Problem: Party Building Puzzles}\label{PBP}
To illustrate our proposed Graph-Based Genetic Algorithm approach  to finding a set of near-optimal solutions we, first, describe an example problem to introduce all the necessary terminology, followed by a constructive search based method to build a single solution, which becomes the basis for the evolutionary computations.

\begin{figure}[t]
\centering
    \begin{subfigure}[t]{0.225\textwidth}
    \centering
      \includegraphics[height=0.8in]{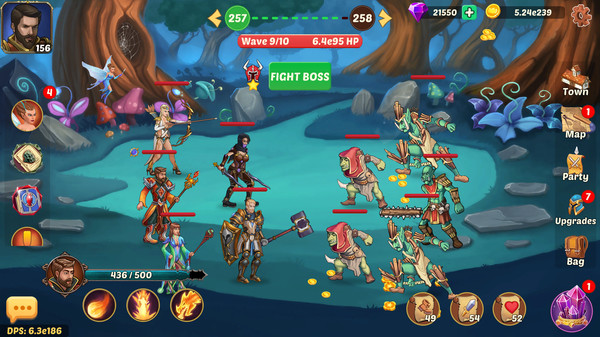} 
         \label{fig:subim1}
         \caption{Idle Firestone}
    \end{subfigure}
    ~
    \begin{subfigure}[t]{0.225\textwidth}
    \centering
        \includegraphics[height=0.8in]{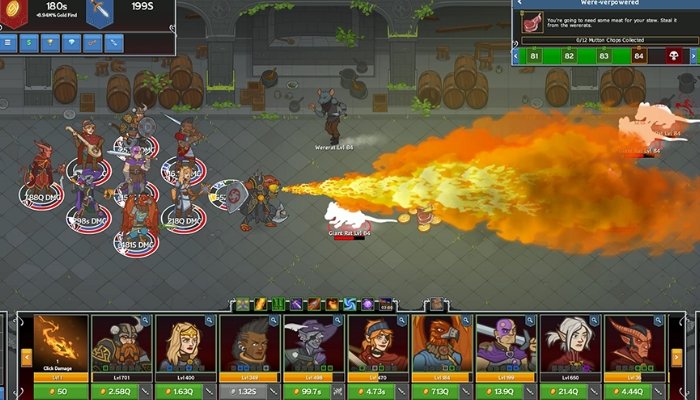}
    \label{fig:subim2}
    \caption{Idle Champions}
    \end{subfigure}
\caption{These figures illustrate fantasy parties arranged in typical combat formations. For example, melee fighters are positioned in the front and ranged fighters or support characters are in the back. Even if the core game play differs, these concepts are shared, suggesting the wide applicability of formation-building puzzles within this game genre.}
\label{fig:image2}
\end{figure}

Party building fantasy combat games are rising in popularity recently, e.g. Idle Champions, Firestone Idle RPG, or others (Fig.~\ref{fig:image2}). 
In these games, a player is tasked with selecting and filling out a combat formation with different heroes. Each combatant has different properties, making them suitable for different positions in the combat formation as well as making them strategically preferable to combine with other fighters. The genre of party-based, fantasy combat games is a suitable candidate for demonstrating our approach.
The game we are analyzing is called ``Party Building Puzzle'', or PBP for short. In a PBP, the player has to select a number of items (heroes), and then assign those items to the combat formation such that heroes in the formation meet the puzzle requirements. An important thing to note is the puzzle requirements are defined by the game designers, which differentiates this type of formation filling from a more general strategy based games, and hence the use of the term `puzzle'. 

Each hero has a number of properties such as race, religion, nation,  total level, gear quality level, price and how rare it is to obtain that particular hero. Puzzle requirements are constraints the player needs to honor when building their formation, and are usually related to the hero properties. For example, a puzzle might require the player use at least 3 orcs in their party, from at least 4 different nations, while another puzzle requires that the average level of the party be at least 60. Heroes additionally contribute to something called party synergy, a graph-based metric of how well the party works together and will be described in more detail in the next section. 
Requesting players to form a party of a certain synergy level creates a complex challenge in which the player must not only choose the right heroes, but also arrange them appropriately within the party's formation.

Currently the game possess a total amount of unique cards in the order of $10^{5}$, and the number keeps expanding as the designers are constantly releasing new cards. The number of possible heroes is near limitless, which creates a combinatorially complex search space over which a solver must iterate to find solutions that satisfy the requirements. 
In that scenario, a task of constructing all possible solutions for a given puzzle becomes infeasable. For such, we use the optimization heuristics of evolutionary strategy to efficiently search the space of possible solutions, while optimizing for what is defined as the ``cheapest' priced option.

\section{Puzzle Solver Part I: Formation Building and Playtesting}\label{sec:formationplaytesting} 
\subsection{Problem Formulation}
The puzzles we are considering, with both linear and quadratic constraints, can be formulated as an asymmetric quadratic assignment problem (QAP), which consists of selecting $N$ items out of a pool of $M$, where $N<<M$, and placing them into a graph of locations in an optimal way such that certain conditions are satisfied. 
Each item, i.e. hero, is characterized by a list of properties, so-called traits. Each can be selected once and its selection and corresponding placement are advised by those traits. In our example of the Party Building Puzzle,  $N=10$ is the number of nodes in the puzzle graph to fill out and $M=10 000$ is the number of possible items to choose from without repetition. 

Linear constraints are refereed as simple, and can be either continuous or discrete. For continuous properties, constraints are formulated as `the accumulated quality of the property $P$ over all positions has to be greater or equal to some value $a$': $\sum_{i=1}^N x^P_i\geq a$. For example, \textit{minimum team level is 84}. Discrete property constraints require `not more than $b$ items with a property $P$ are allowed in the placement graph': $ \sum_{i=1}^N I(x_i^P)\leq b,$ where $I$ is a boolean indicator function defining the presence or absence of a certain trait in the feature vector for each item. For example, \textit{at most 8 elves are allowed} or \textit{at most 4 humans and at least 2 elves are allowed}. In our PBP, number of properties for a hero does not exceed $P=8$.

In addition, we have a nonlinear  complex constraint, so-called `synergy',  which is dependent on both item placement and the compatibility  between adjacent items in the graph. Hence, its value is not  defined till all the items are selected and assigned. Synergy is present within each puzzle, but can have various requirement value from 0 to 1. 
\begin{equation}
    \label{complexconstraints}
\sum_{j=1, j\neq i}^N \sum_{i=1}^N \sum_{p=1}^Pw(x_j^p, x_i^p) \geq Synergy,    
\end{equation}
where $w(x_p^j, x_p^i)$ is the weight of the edge in the graph between two neighbouring nodes $i$ and $j$, and summations are for each edge, across each trait.  The weight function is specific to the puzzle and is a custom-defined, non-linear relation.

We formulate Quadratic Optimization problem to maximize the synergy constraint following the Koopmans and Beckmann QAP formulation~\cite{QAPoriginal} over all possible selection and assignment permutations. Our solution must search for which items to use, out of the available ones, as well as which position to place the items in. 

In addition to the classical QAP, 
we need to also pre-select $N$ out of $M$ possible items before assigning them to locations, which makes the assignment asymmetrical. 
The asymmetric nature of the placement presents unique challenges, since the number of candidate items $N$ can be on the order of $10^{5}$, while the graph consists at most a few dozen items. So, in addition to the optimal placement, the selection of items is part of the optimization.

\subsection{Our Proposed Solution}\label{sec:approach} 
In this section we first describe the design to automatically build a deck by finding a feasible solution to a challenge, i.e. the one that honors all requirements specific to that puzzle while maximizing the synergy. Using this design, we further employ evolutionary approach to finding a set of optimal solutions as described in the following section. 

It is known that QAPs  are difficult to solve and are among the hardest NP-complete problems~\cite{puzzlenpcomplete} especially given the size of the combinatorial search space.  Any algorithm that guarantees an exact solution (given that it exists) has to consider every item (hero) and every combination, so has an exponential computational time. Heuristic driven approaches are usually employed to either linearize or find approximate solutions~\cite{HandbookMetaheruistics}.

Potentially, training a Reinforcement Learning agent to solving a puzzle could lead to a fast performance. However, with dynamically changing sets of requirements as well as a large pool of candidate items it could lead to overfitting and overall would result in data inefficiency in training and might not be practical. There are additional complications of defining an adaptable state representation, as well as  catastrophic forgetting (both during training and/or during re-training with new data).  This is due to the fact that, though there is a limited, constant, set of well-defined requirements that may appear in a given puzzle, each puzzle has a unique combination of completion requirements. Compare to Sudoku solvers~\cite{sudoku2018} as an example, where similar graph type constraints are defined, but the rules of the game are always the same even when the initial state of the board is different for each puzzle. In addition, placing numbers from 0 to 9 is not comparable to selecting 10 items out of tens of thousands and then placing them optimally. 

Another set of approaches commonly applied for similar problems are  classical search methods like A-star or Monte-Carlo tree search that are based on heuristic-driven or probability-based backtracking~\cite{brownie2012MTCSreview}. However, for problems with quadratic constraints like the graph-dependent synergy value that we are trying to maximize, those approaches have limited application.  In particular, deriving  admissible heuristics  for  quadratic optimization is a challenge by itself. Additionally, the size of the search space we are dealing with makes those approaches impractical.

As a reminder, our goal is to select $N$ items, each of which has $P$ features, from a list of $M$ items, where $N<<M$. These items must be selected and assigned to positions within a graph so as to satisfy a list of requirements and avoid repetitions.
The solution strategy we chose is a constructive, randomized, heuristic search as described in Algorithm~\ref{algo:constructsolution}. To avoid deterministic behaviour, we first define a random traversal order in which to visit each node in the graph. Then, for any unvisited node, we first filter out all items from the list of candidates  following the constraint requirements such that any item chosen from the filtered list is valid. Once the list is filtered, we search through it to find an item that maximizes the synergy of the neighbouring candidates that are partially filled. The random path traversal ensures diversity, so that the process can be repeated until the synergy requirement is satisfied.

\subsubsection{Filter the candidate pool}
Prior to selecting items at each position that optimize for the synergy, at each step of the algorithm iteration we apply heuristics specific to the problem and the set of requirements in order to filter out all the candidate items that are not eligible for the valid assignment. This look-ahead approach allows to do forward checking between the current and future candidates and if at any point the candidate pool is empty given an early signal to start over. This step is applied sequentially $N$ times for each linear constraint  for each position until each node in the graph is visited. 
This heuristic rule is applied for discrete and continuous feature constraints.

Consider a given step of solution building process when $L$ positions out of $N$ are already assigned (or, equivalently, visited), $0\leq L<N$, and at least $a$ items of a specific property $P$ are required: $\sum_{i=1}^N I(x^P_i) \geq a$. 
First, we compute the intermediate value  $l$ of that constraint using assigned items: $l = \sum_{i=1}^L I(x^P_i)$. Then,  $l \geq a$ means that the constraint is already satisfied and no action is taken. Alternatively, if $l < a$, we have  $K=N-L$ nodes yet to be visited, out of which remaining $k=a-l$ items have to have a property $P$. Then, if $K>k$, again no action is taken. Else, if $K=k$, we filter out all items in our candidate pool that do not have property $P$ forcing in subsequent iterations the selection of a property $P$ to honor the requirements. This process is repeated for each requirement. Note, that since at each iteration only one item is selected to fill-up a given node, the repetitive application of these heuristics guarantees that $K$ is never less than $k$, and at the end all requirements are satisfied.

\subsubsection{Select the best matching item}
\begin{figure}[b]
\centerline{\includegraphics[scale=0.4]{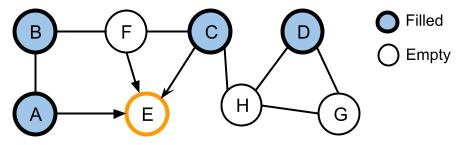}}
\caption{Graph traversal for searching for the best matching item based on synergy with neighbors. The order of traversal is alphabetical, so that node $E$ is influenced only by nodes $A$ and $C$, since they were the only neighbours visited.}
\label{graphBestMatching}
\end{figure}

Once the item candidate pool is filtered, any randomly selected item will satisfy all the linear constraints. At this step the goal is to select the best matching candidate in the partially filled graph which maximizes complex constraints. The complex constraints are computed for a position that has to be filled with respect to its immediate neighbors in a partially filled graph. 
For example, if the node $E$ has to be filled as shown on Fig.~\ref{graphBestMatching}, it is only influenced by nodes $A$ and $C$. For an empty graph, a random item is picked. 

\begin{figure}[h]
    \centering
    \begin{subfigure}{0.235\textwidth}
      \includegraphics[height=1in]{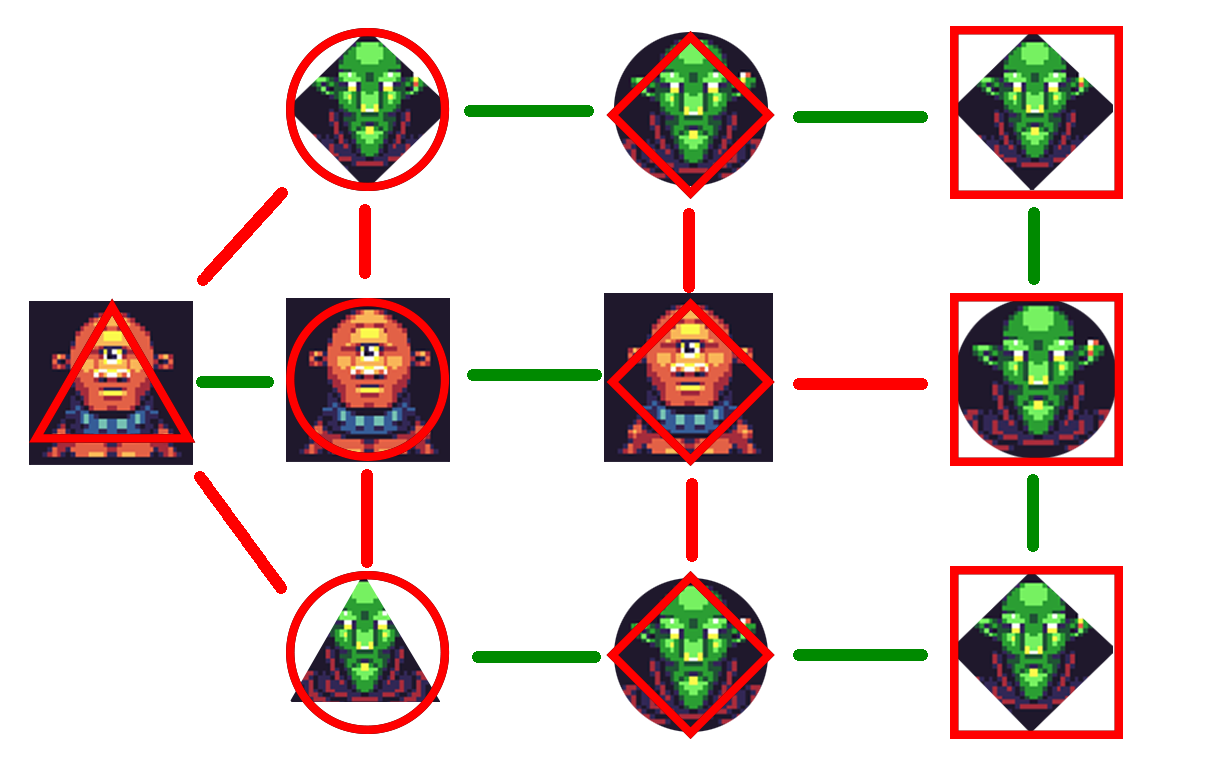} 
    \end{subfigure}
    \begin{subfigure}{0.235\textwidth}
        \includegraphics[height=1in]{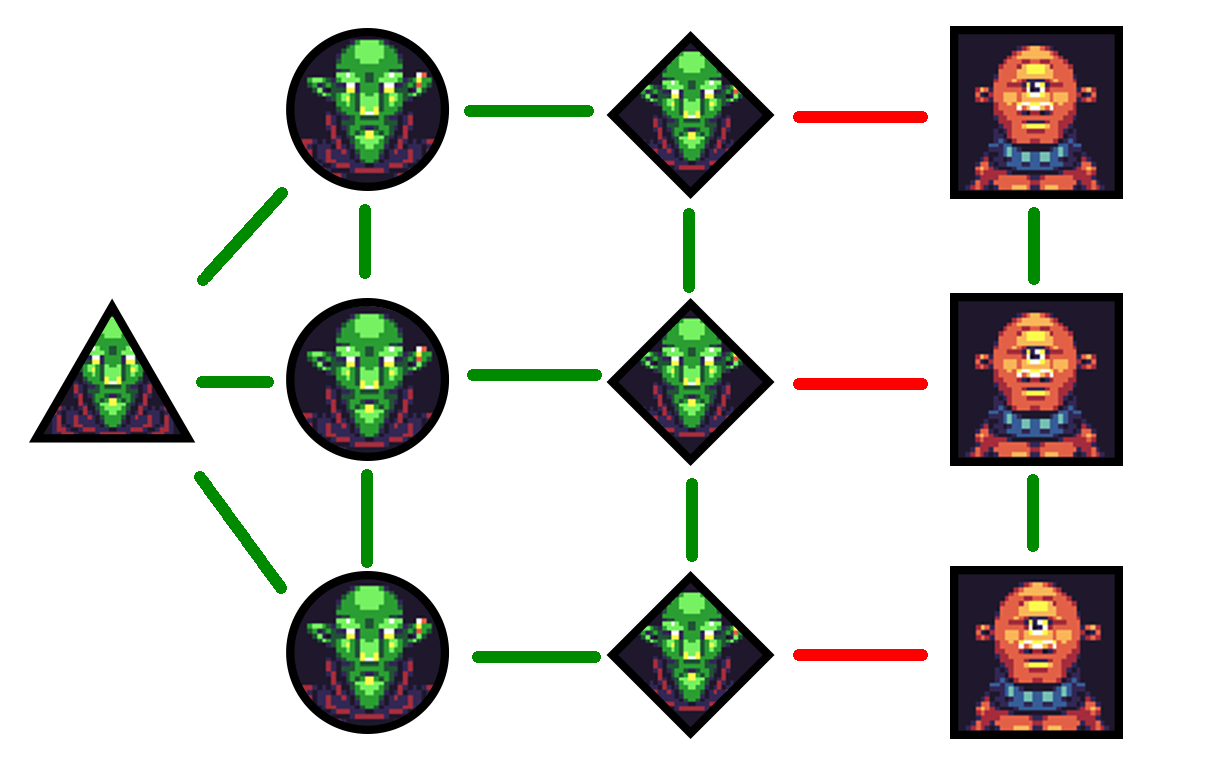}
    \end{subfigure}
\caption{Sample solutions to a puzzle with the requirements `At least 7 Goblins and at least 2 races.'. In both cases the same heroes are used. However, the arrangements are different which result in a higher synergy (edges) for the solution on the right as party members are placed more optimally.}
 \label{fig:samplesolutions}
\end{figure}

\begin{figure}[h]
\centerline{\includegraphics[width=0.7\linewidth]{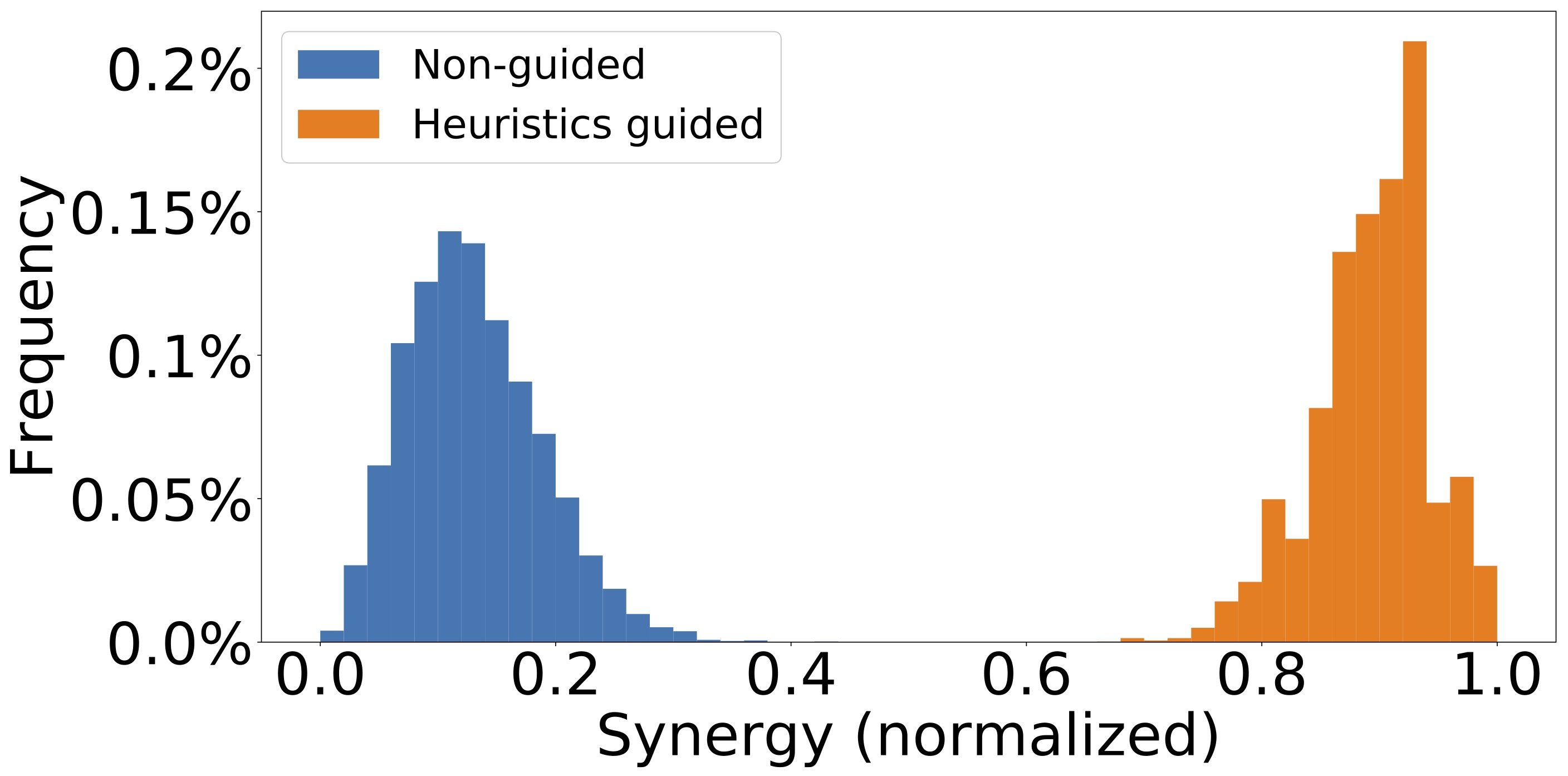}}
\caption{Comparison of Synergy values normalized to [0,1] for an independent set of 5000 attempts for a given challenge with and without heuristics guidance. Vertical axes are frequencies as  percentage over all the samples. As can be seen, without the heuristics guidance to continuously select the best matching items, the synergy value never reaches more than 0.3. Results are for a sample challenge.}
\label{fig:chemistry_unguided}
\end{figure}

Of note is the fact that this intermediate synergy value at any given step does not guarantee that the final solution will comply with the complex requirements since these are quadratic requirements. If they should fail to be satisfied, the algorithm starts over with a new randomly selected traversal path.  Fig.~\ref{fig:samplesolutions} demonstrates two sample solutions which satisfy all linear constraints, which concerns the selection of the items, but only the right one satisfies quadratic constraint (both selection and optimal arrangement). Since the synergy is, however, usually hard to obtain, requiring a number of attempts for a given search space which makes the problem intractable, our proposed guided, randomized search increases the chances of obtaining feasible solutions in only a few iterations.
Fig.~\ref{fig:chemistry_unguided} shows the results of running the algorithm on a given set of linear requirements with and without heuristics guiding it toward synergy maximization. In both cases same constructive approach is used to build a solution, so linear constraints are satisfied through the candidate filtering, the only difference is that we do not select the best matching candidate in the later case.   Out of 5000 independent attempts, the maximum synergy value reached without heuristic-guidance was 0.35. This is poor performance, as common in game required values are typically above 0.7-0.8, demonstrating the value of heuristic-guidance.
Solution construction is summarized in Algorithm~\ref{algo:constructsolution}.

\begin{algorithm}[h]
\SetAlgoLined
\begin{algorithmic}[0]
\State{\bf Initialize}: Random graph traversal path

\While{Synergy is not satisfied}{
    \For{Each unvisited node in a graph}{
        \State Look ahead: filter the list of candidate items
        \For{all linear requirements}{
            Apply the heuristics rule
            }
        \State Select the best matching candidate  (synergy wise)

  }
   \State Compute synergy Eq.~\ref{complexconstraints}

   \If{synergy requirement is satisfied}{return the solution: {selected items $x$, permutation $\sigma$}}

   \Else{
   New random path and iterate until solution is obtained}
 }
\caption{Guided Randomized Heuristic Search to Construct a Solution}
\label{algo:constructsolution}
\end{algorithmic}
\end{algorithm}

The computational complexity of the algorithm is ${O}(kNM)= {O}(NM)$ per iteration, in the worst-case, where $k$ is the fixed finite number of linear requirements on the order of $1$ to $6$, $N$ is $10$, the number of nodes in the graph, and $M$ is the size of the candidate pool, order of $10^4$. The number of iterations is fixed, typically to a value of $10$ or less for the purposes of our experiments. In the best case scenario, however, linear requirements lead to a significant reduction in the size of the candidate pool such that it approaches the size of the graph. In these cases, the complexity approximately reduces to ${O}(N^2)$ per iteration. 

Important to note that exhaustive search is intractable for this type of a  problem, since the number of all possible combinations of (a) selecting 10 items out of 10 000 and (b) arranging those 10 items into a specific order is roughly of the order of $10^{39}$. Depending on the requirements of the puzzles, there could be a few hundred thousand possible feasible solutions out of those combinations.

At the same time, human players are able to find solutions without extensively covering the entire search space by using a limited number of items available to them and obtaining a few missing ones, as well as taking advantage of their intuitive knowledge of the game mechanics and items’ properties. Note, the heuristics used in the algorithm are inspired by how human players act: the way they select items and use in-game tools to filter out certain items’s traits, and how they use their intuition to arrange the items to increase synergy.  However, the task of a game designer to find not just a solution, but the ‘best’ solution requires extensive coverage of the entire search space, not just using those few items available to each player (and which could of course be different for each player). 
 
\section{Puzzle Solver Part 2: GA-Based Solution Optimizer}
\label{sec:optimization}
The solution space of the type of puzzles described often times consists of multiple solutions, each which are valid, honoring the constraints, however some solutions are more desirable than others, like having a low price or using items which are easier to obtain. Knowing the optimal solution allows content creators to gauge the desired reward value and puzzle difficulty accordingly. 

Optimizing for a solution in the class of constraint satisfaction problems considered poses unique challenges. Since they form discrete combinatorial problems, gradient based optimization methods are not applicable. Various population based optimization methods have been applied to solving similar type of puzzles.
In many cases, memetic variants of genetic algorithms, i.e. evolutionary approaches hybridized with constraint satisfaction tools,  are used successfully for combinatorial optimization problems~\cite{HandbookMetaheruistics}.
Similar ideas were also used as a hybrid genetic algorithm, such as when applied to the Light-up puzzle~\cite{LightupPuzzlesHybridGA2007}. In those approaches, the genetic algorithms always work with feasible solutions, and if the individuals become infeasible after crossover and mutation operations, they are being ``healed'' to restore feasibility. 
In this work we propose a customized hybrid GA algorithm for finding the optimal solution in terms of an objective not part of the constraint by searching the solution space. 

\subsection{Graph-Based Hybrid Genetic Algorithm for Solution Optimization}
There are various ways in which we can choose to setup the architecture of the genetic algorithm, such as representing the solution space, deciding to allow vertical or horizontal crossover, which of multiple ways parents are selected for crossover based on their fitness value, which mutation rate, mutation strategy, and selection strategy to use, and the settings for initial population size, offspring size, crossover and mutation rates. We are using a Graph-Based Genetic Algorithm (GB-GA) in which the relative locations of nodes to be filled are important, so reshuffling of genes is not allowed~\cite{Ashlock1999GraphBased}. The algorithm steps are described below highlighting the specifics of our implementation.

\subsubsection{Representation and Initialization}
 In our terminology, a chromosome is an individual solution to the puzzle, which consists of selected items from the candidate pool assigned to the formation. Same items assigned in a different order are considered as two separate solutions. To start off a genetic algorithm, Algorithm~\ref{algo:constructsolution} is used to randomly generate a number of independent feasible candidate solutions, i.e. chromosomes.
\subsubsection{Crossover and Mutation} 
After examining several variants of crossover, we have chosen a uniform crossover operation such as each unoccupied position in the offspring solution (chromosome) is assigned an item (gene) from one of the parent solutions with the probability $p=1/2$, while enforcing that no item may be repeated in the offspring solution. The selection of parents for the crossover is performed by a rank-based selection rule~\cite{TATE1995}. 
The mutation rate is set at 20\%, meaning that each individual has a 20\% chance being replaced, resulting in an average of $0.2N$ items removed from the graph during a given mutation pass. In detail, a solution is traversed in a random order and at every step a probabilistic decision is made whether to remove a given  item both from the solution and the remainder of the candidate pool. The resultant partial solution is then populated again where missing items are randomly filled with the new items (heroes).  

\subsubsection{Hybrid Approach to Find Feasible Solutions}
Note, that solutions produced by recombination and mutation do  not necessarily  result in feasible ones, since those operations are not guaranteed to satisfy the challenge constraints. As an extra step before accepting them, we repair those solutions by a process we call `healing'  to  ensure that the final offspring solution is feasible, or valid~\cite{LightupPuzzlesHybridGA2007, HybridGA2}.

More specifically, first, those few items in the offspring that were replaced during mutation are removed. Then, the same Algorithm~\ref{algo:constructsolution} we used to build a new solution is used to replace missing items on the partial solution. If no such replacements are possible, the algorithm rejects that solution in favor of a new randomly generated one.

\subsubsection{Selection and Refreshment}
The selection process favors solutions with better fitness and diversity. The new generation of solutions are selected from the offspring and previous population. If the selected solutions comprise a diverse enough population, the algorithm moves to the next generation by selecting the best fit individuals. Otherwise, if diversity drops below a threshold, a fixed number of new randomly generated individuals are added to the offspring pool.


\subsection{Maintaining Diversity}
One of the key reasons the algorithm encounters premature stagnation, or trapping, is when the population looses diversity among its individual solutions. If that happens, recombinations and local perturbations through mutations are not able to lead the individual solutions to escape local minima~\cite{Whitley1995}. We have chosen phenotype diversity as a representative measure relating to the optimal fitness of the solution.

We  explicitly track diversity within population at every generation, and if it falls below a certain user defined threshold (measured by a normalized coefficient of variation),  we randomly replace a third of the existing solutions with the new individuals\footnote{Both the diversity measure and threshold as well as the percentage of the solutions to be replaced are engineering hyperparameters that have been tested to be efficient in average.}. While maintaining diversity  is essential to avoid algorithm stagnation, the diversity measure is still a proxy as it does not capture full variations in the solution space. In addition, adding new solutions that are far from the optimal reduces the rate of the convergence. Alternatively, one can introduce an adaptive mutation rate, which increases the number of mutations when the diversity of the population stagnates. These approaches, however, slow down the generation process. Additionally, mutation does not always lead out of a local minimum.

One of the most successful solutions to this problem is the ``multi-population'' or ``multi-island'' model~\cite{Leito2015IslandMF, Tomassini2005, Whitley1995}, which allows a unique search trajectory to be followed by each islands, resulting in a more efficient exploration of the search space. An additional benefit is that it is inherently parallelizable and can be implemented employing distributed computing.

In our current work we have designed a multi-island approach with migration to encourage the genetic process  in maintaining diversity.  The overall population is partitioned into sub-populations based on their similarity and each sub-population is assigned to an island. After that, each island evolves independently, and then, after a fixed number of sub-generations or epochs, migration allows the islands to interact. 
The traditional island model requires additional parameters such as the number of islands, size of the population of each island, the migration frequency, migration rate,  migration topology and migration policy. The migration strategy between the islands directly affects the performance. It also impacts the optimization of the algorithm via balancing exploration and exploitation and indirectly supporting diversity. As a result, each island explores a separate optimization path resulting in broader coverage of the search space. 

The exact mechanism of migration between islands used in this paper is following a fully-connected migration pattern: after a fixed number of sub-generations within islands (10 sub-generations), all solutions across all islands are combined, sorted out by their fitness similarity, and then divided amongst  islands  equally such that most similar solutions are grouped together.

There are multiple choices of the migration strategies~\cite{Li2015HistoryBasedTS} since islands can be clustered together using various similarity measures, either according to their fitness values, or through other measures of diversity like entropy~\cite{Arellano-Verdejo2017}, or even dynamically through spectral clustering based on the  pair-wise similarities between individuals~\cite{Meng2017}. 
In general, having more densely connected islands gives a higher accuracy of the lower bound, but it is more computationally expensive. In the current work we have chosen the fully-connected island model where migration of individuals is not constrained and to use fitness as our similarity measure.

\subsection{Experimental Setup}
\begin{figure*}[t!]
   \centering
   \subfloat
     {\includegraphics[width=0.31\linewidth]{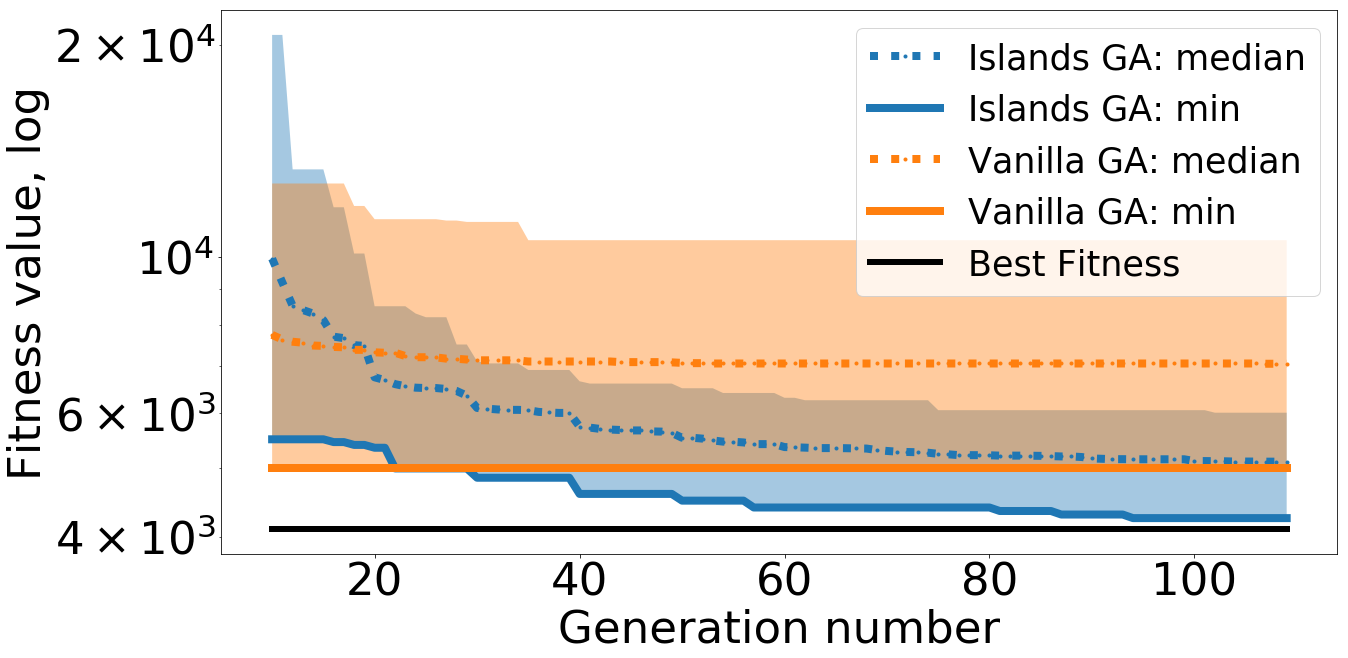}}
   \subfloat
     {\includegraphics[width=0.31\linewidth]{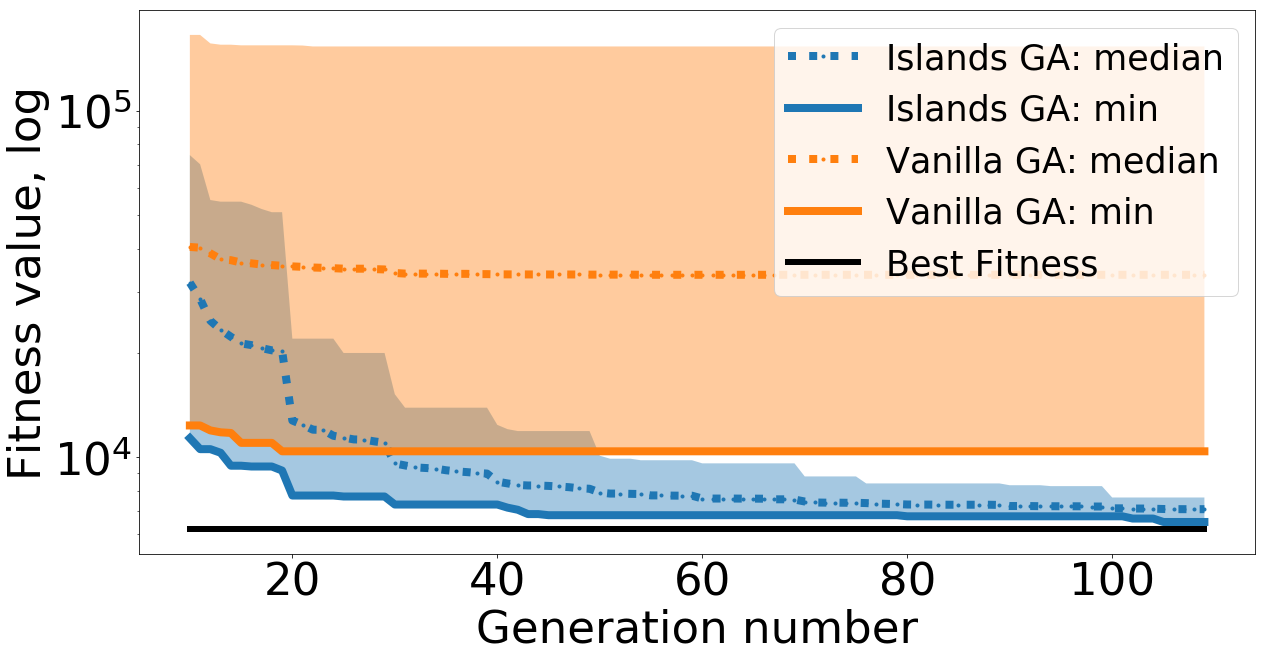}}
   \subfloat
     {\includegraphics[width=0.31\linewidth]{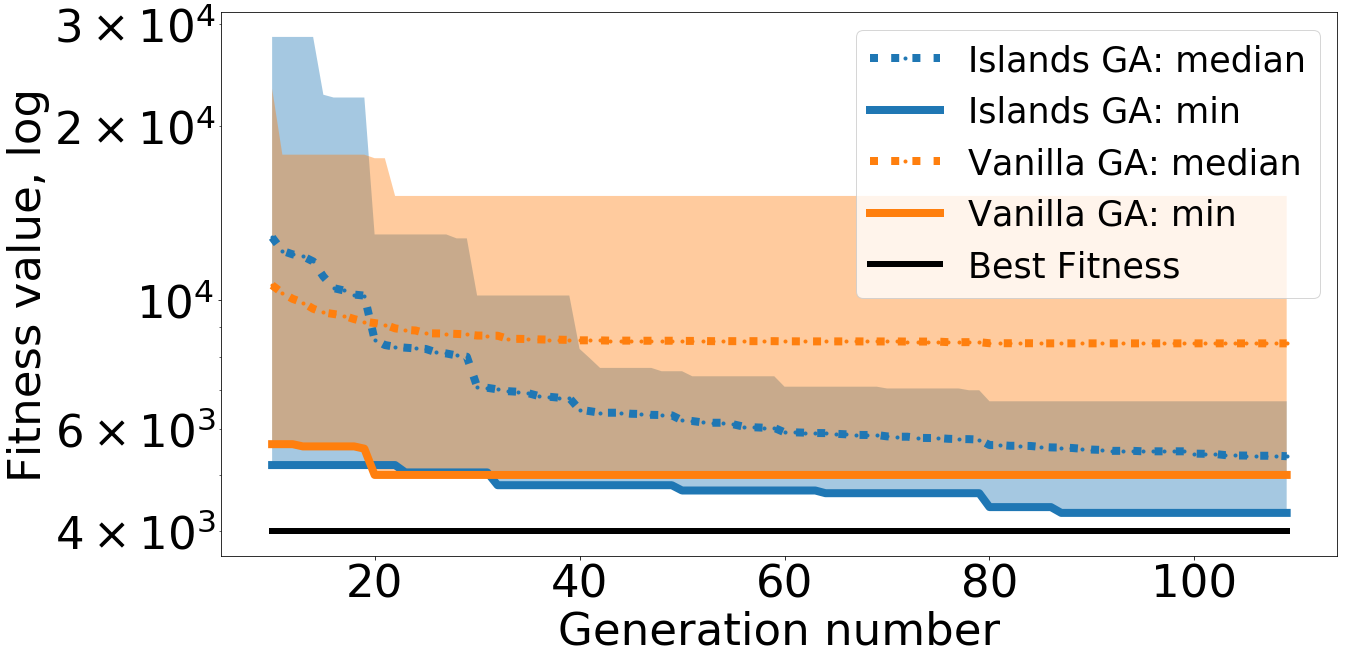}} \\
   \subfloat
     {\includegraphics[width=0.31\linewidth]{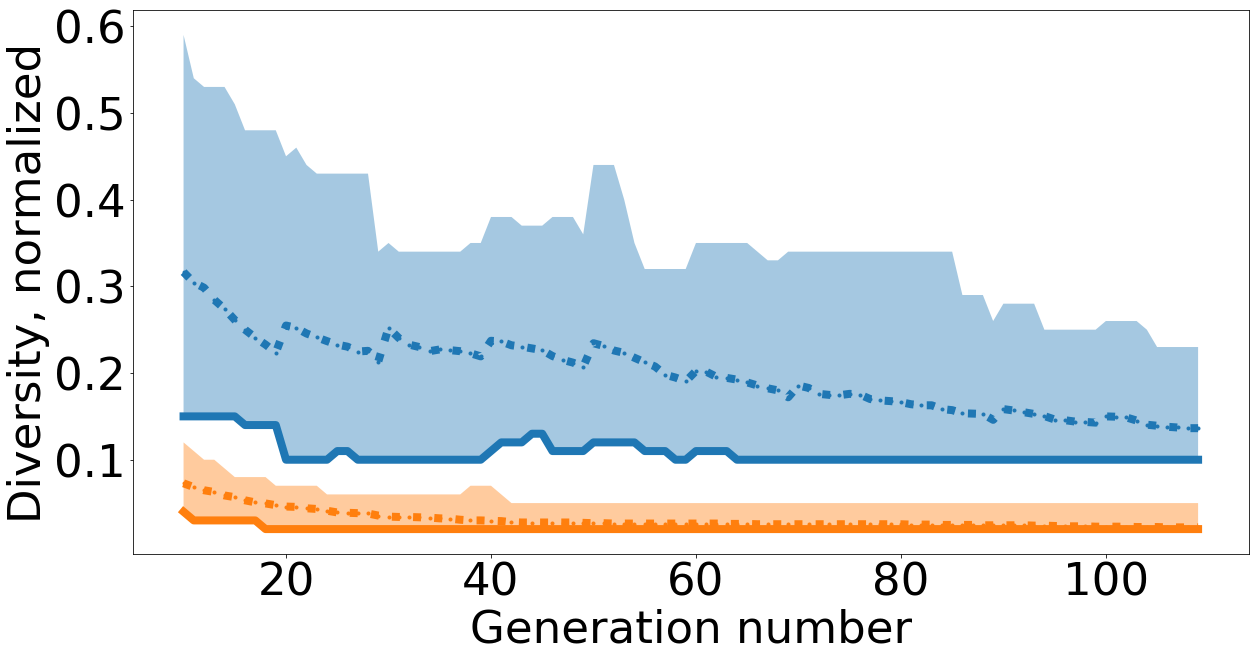}}
   \subfloat
     {\includegraphics[width=0.31\linewidth]{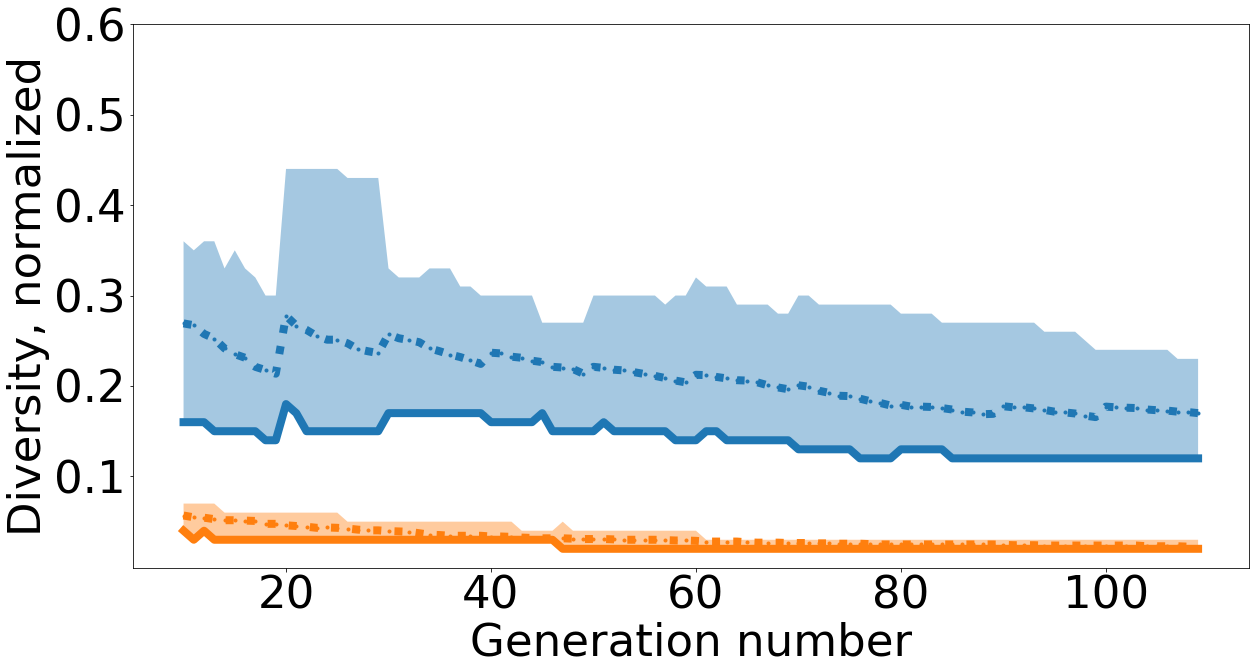}}
   \subfloat
     {\includegraphics[width=0.31\linewidth]{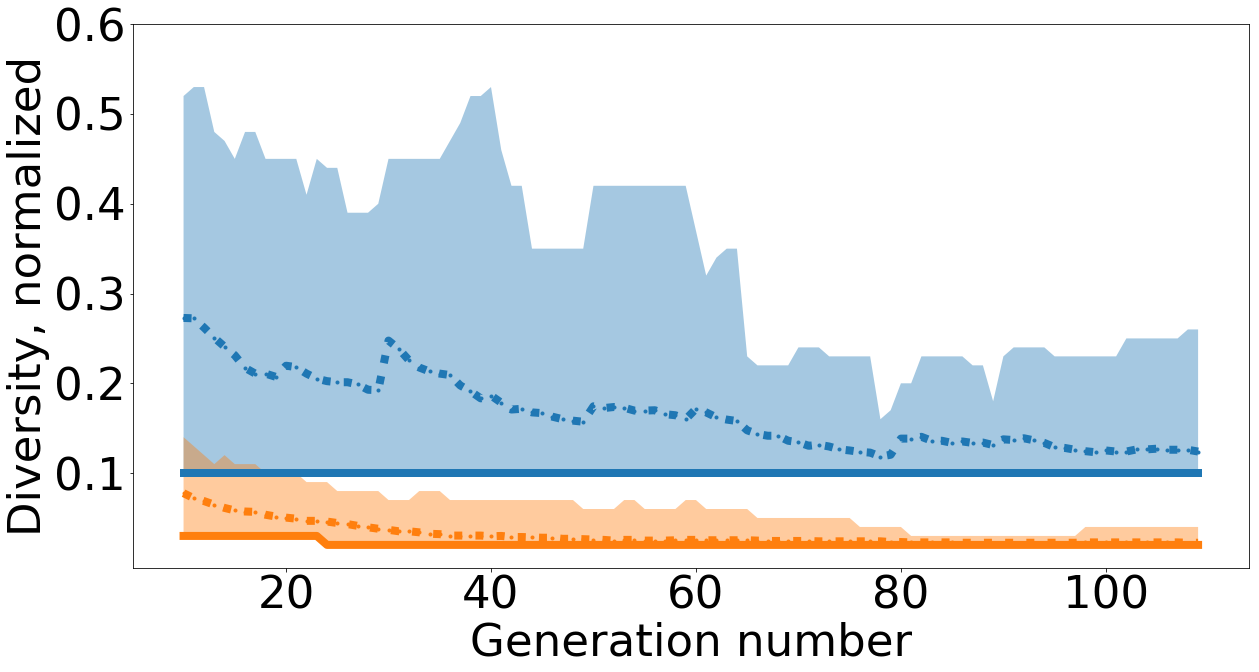}}
  \caption{\textbf{Top row}: Convergence curves of best fit solutions for challenges Type1, Type2 and Type3, comparing the two approaches while maintaining diversity in both cases, averaged over 50 independent runs. The solid lines indicate the best fitness value across all the runs, the dotted lines are median values at each generation, and shaded regions cover each run. The solid black line represents the optimal solution. \textbf{Bottom row}: Diversity at each generation, solid lines: minimum, dotted line: median.}
  \label{fig:convergence}
\end{figure*}
To demonstrate the performance of our approach, we defined 3 types of puzzles that cover different requirements and sizes of the search space. Certain requirements increase the difficulty of achieving a specific synergy threshold, if they create a large search space. Opposed to such, are constraints that greatly reduce the number of candidate items, resulting in quicker convergence, since synergy is easier to satisfy with more similar items. Our puzzles representative of each puzzle types are:

\begin{itemize}
    \setlength\itemsep{0.05em}
    \item{\textbf{Type 1}} Min Synergy 0.8, Min Team Level 84
    \item{\textbf{Type 2}} Min Synergy 0.9, Min Team Level 75, at least 8 religions, at least 6 hometowns, at most 2 heroes with the same religion
    \item{\textbf{Type 3}} Min Synergy 0.8, Min Team Level 84, Min of 8 elves
\end{itemize}

For each of the selected challenges the optimal solutions are provided by the game designers as benchmarks. The parameters of the Genetic Approach were selected through trial and error to perform reasonably well across various puzzles as summarized in Table~\ref{TableParameters}. 
Our goal was to demonstrate the advantage of multi-islands approach in maintaining diversity, in comparison with a single population approach, and thus leading to more reliable results, while the actual  model parameters can be tuned for a problem of interest. For example, authors in~\cite{RollingHorizon2020} present an N-Tuple Bandit Evolutionary approach to automatically optimize hyperparameters.
\begin{table}{}
\begin{center}
\caption{GA parameters for ``vanilla'' and ``multi-islands''}
\begin{tabular}{lll}
\hline 
Parameter & vanilla GA & multi-island GA \\
\hline
 mutation rate & $0.2N$ &  $0.2N$ \\
   crossover rate & 0.5 & 0.5 \\
  pop/offspring size & 50/100  & 10/20 per island \\ 
  migration type & N/A & similarity based\\
  similarity measure & N/A & fitness \\
  migration frequency  & N/A & 10 \\
  islands epochs & N/A & 10 \\
  generations & 100 & 100 \\ 
\hline
\end{tabular}
\label{TableParameters}
\end{center}
\end{table}

\subsection{Numerical Results}
The experiments were conducted for the three types of puzzles, comparing ``vanilla'' hybrid genetic algorithm with the ``multi-island'' approach.  Fig.~\ref{fig:convergence} compares the behaviour of both approaches, the difference between those two approaches across 25 independent restarts  with a fixed computational budget for each scenario so that average performance can be traced.  Performance is defined by the rate of convergence and solution population diversity.  
During the early generations,  the ``vanilla'' genetic approach slightly outperforms the multi-islands since it has more individuals to choose from.  While average over time, shows the multi-island approach better approximates the lower bound with less variance as it manages to escape the local minima by interaction between the islands. We conclude that overall multi-islands approach consistently outperforms the standard GA under the same time, while maintaining at least twice higher diversity. The effect of the initial generations fades out fast for both approaches.  The steps on the convergence curve reflects the reshuffling between the islands.

\subsection{Notes on Algorithm Feasibility}
As with any Genetic Algorithm for combinatorial non-convex optimization, there is no way to prove that any of the local minima are actually the global minima. The approach we took is to show that even though the algorithm is random in nature (both because of initial population and recombinations), if we run it repetitively for the same puzzle we could show that initial population, as well as a series of random crossover/mutations, are all leading to the same minimal fitness value solution, in average. Even more, by maintaining the diversity of the population through a multi-island approach, we could avoid algorithm stagnation and keep exploring the solution space as long as a satisfying minimal solution is found.

In theory, the actual global minima is not known for these types of problems. So we are effectively comparing our genetic approach to an alternative of finding and enumerating all possible solutions and selecting the `best' one. That could be possibly finding a few hundred thousand solutions. Even though each of them takes about 10 sec in average to find on a standard cpu machine, it would overall take up to $300$ hours ($10 \times 100 000 / 3600$), while with the Genetic Algorithm we can limit that time to up to $10$ minutes, by reducing the number of independent solutions we need to find to only initialize the population, while recombinations are relatively cheap to compute. The number of puzzles generated by designers then becomes an important factor, as they are constantly generating new puzzles (on average $20$ per day) and would need the solver to run in actionable time to gain the knowledge of the attributes of the solution space.

\section{Conclusion and Future Work}\label{sec:conclusionfuture}
Our approach is motivated by the recurring problem designers have of improving and optimizing the task of creating new quality puzzle variations. We demonstrate our use case on the Party Building Puzzle game, where players collect heroes and later select from their collection which ones to use and how to arrange them in order to complete puzzle constraints. Our approach optimized for minimum amount of in-game resource value our solution has, to help designers evaluate and compare the puzzles they create. In addition, the proposed solution had to be capable of running in feasible time to allow designers to constantly iterate over their designs in just a few hours.

In this work we have proposed an efficient constructive randomized search algorithm to build a solution using heuristics specific to the puzzle's constraints, and show that a solver hybrid, graph-based, genetic approach allows us to find near-optimal solutions to the puzzles. One of the challenges of these types of combinatorial optimization is an early drop into the local optima. We have experimentally demonstrated how the multi-island approach with a randomized selection strategy allows us to reach a near-optimal solution. 

As mentioned above, instead of decoupling the constraint optimization problem into two different ones: constraint satisfaction for a QAP and a combinatorial optimization to find the best performing solution, one could alternatively formulate the problem as a  multi objective optimization, where constraints like synergy, ratings etc are optimized in combination with the fitness. This exploration of a comparative performances of these approaches is a future work.

An avenue to explore are methods that can further investigate the search space of solutions. In an effort to increase diversity of potential solutions, and thus provide designers with a more detailed insight into their own design, we plan to explore algorithms for illuminating search spaces, such as Map-Elites~\cite{khalifa2018talakat, fontaine2019mapping}. Map-Elites explores different dimensions of the search space, and with such also provides an alternative solution that is more robust in avoid the local minimal problem discussed.

We have demonstrated that power of Genetic Algorithms could be exploited for the NP hard combinatorial optimization problem with large non-unique search space and in the presence of additional constrains. Our proposed framework with the novel puzzle representation and custom designed multi-islands graph based genetic approach  could be adapted to other problems with similar properties as long as the genetic operations are specified for the problems of interest. This method could also further be extended to solve puzzles starting from a partial state since there is no dependency of prior state.

\section*{Acknowledgment}
We would like to genuinely thank the anonymous Reviewers for all valuable comments and suggestions, which helped us to improve the quality of the manuscript.

\bibliographystyle{IEEEtran}
\bibliography{aiide_refs.bib}
\end{document}